%% file: main.tex
\PassOptionsToPackage{table}{xcolor}
\documentclass{article}
\usepackage{latexml}
\iflatexml
\usepackage{times,natbib}
\else
\usepackage{iclr2027_conference,times}
\fi

\usepackage[T1]{fontenc}
\usepackage{amsmath,amssymb}
\usepackage{graphicx}
\makeatletter
\newcommand{\includevectorpdf}[2][]{%
  \begingroup
  \@ifundefined{Gread@transgroupfalse}{}{%
    \let\Gread@transgrouptrue\Gread@transgroupfalse}%
  \includegraphics[#1]{#2}%
  \endgroup}
\makeatother
\usepackage{booktabs,array,tabularx}
\usepackage{multirow,makecell}
\usepackage{xcolor}
\usepackage{placeins}
\usepackage{microtype}
\usepackage{hyperref}
\usepackage{xurl}
\usepackage{url}

\hypersetup{hidelinks,
  pdftitle={CoEvoke: Evoking Multi-Agent Collaboration for Trajectory Collection},
  pdfauthor={Lin Wu, Shu Yao, Yuheng Wang, Jingru Fan, Huatao Li, Yihan Li, Liu Yixuan, Cheng Liu, Jian Chang, Zeyu Zhao, Aohua Zhang, Huiying Zhao, Shuyuan Zhu, Chen Qian}}
\newcolumntype{L}[1]{>{\raggedright\arraybackslash}p{#1}}
\newcolumntype{Y}{>{\raggedright\arraybackslash}X}
\newcommand{\code}[1]{{\small\urlstyle{tt}\nolinkurl{#1}}}
\newcommand{\pass}{\textsc{Pass}}
\newcommand{\fail}{\textsc{Fail}}

\definecolor{auditblue}{HTML}{244A73}
\definecolor{coateal}{HTML}{087F83}
\definecolor{coapurple}{HTML}{6554A4}
\definecolor{coaorange}{HTML}{C56B29}
\definecolor{coagray}{HTML}{F3F5F8}
\input{table_style}
\title{CoEvoke: Evoking Multi-Agent\\Collaboration for Trajectory Collection}
\input{institution_logos}
\input{authors}
\iflatexml\else
\iclrfinalcopy
\input{preprint_title}
\fi

\begin{document}
\maketitle
\iflatexml\else
% Replace the conference status supplied by the template with a preprint header.
\fancyhead[L]{Preprint}
\thispagestyle{coevokeTitle}
\fi
\input{sections/abstract}
\input{sections/introduction}
\input{sections/related_work}
\input{sections/method}
\input{sections/dataset}
\input{sections/experiments}
\input{sections/results}
\input{sections/discussion}

% Author statements and references follow the main text in the preprint.
\input{sections/statements}
\bibliography{references}
\bibliographystyle{iclr2027_conference}
\end{document}

%% file: institution_logos.tex
\newcommand{\coevokeSAISymbol}{%
  \raisebox{-1pt}{\includegraphics[width=18.95096pt,height=8pt,alt={SJTU SAI logo}]{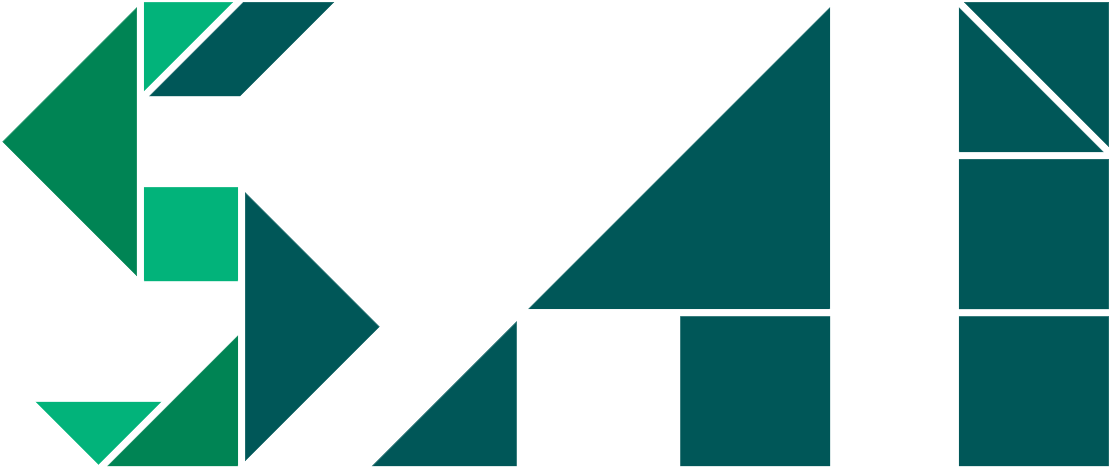}}%
}
\newcommand{\coevokeUnicomSymbol}{%
  \raisebox{-1pt}{\includegraphics[width=11.17377pt,height=8.52pt,alt={China Unicom logo}]{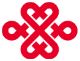}}%
}

%% file: authors.tex
\iflatexml
  \input{authors_html}
\else
  \input{authors_pdf}
\fi

%% file: authors_html.tex
\author{%
\lxAddAnnotation{:cssstyle={width:auto;max-width:calc(100vw - 3rem);min-width:0}}%
\lxWithAnnotation{:cssstyle={width:auto;max-width:calc(100vw - 3rem);min-width:0}}{%
\begin{minipage}{\linewidth}
\centering
\textbf{%
Lin~Wu$^{1}$\quad%
Shu~Yao$^{1}$\quad%
Yuheng~Wang$^{1}$\quad%
Jingru~Fan$^{1}$\quad%
Huatao~Li$^{1}$\quad%
Yihan~Li$^{1}$\quad%
Liu~Yixuan$^{1}$\quad%
Cheng~Liu$^{1}$\quad%
Jian~Chang$^{2}$\quad%
Zeyu~Zhao$^{2}$\quad%
Aohua~Zhang$^{2}$\quad%
Huiying~Zhao$^{2}$\quad%
Shuyuan~Zhu$^{2}$\quad%
Chen~Qian$^{1}$}\\
\mbox{\includegraphics[alt={SJTU SAI logo}]{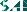}}\quad $^{1}$ Shanghai Jiao Tong University\\
\mbox{\includegraphics[alt={China Unicom logo}]{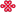}}\quad $^{2}$ China Unicom Online Information Technology CO.,Ltd\\
\href{mailto:linwuai2026@gmail.com}{\texttt{linwuai2026@gmail.com}}\quad
\href{mailto:qianc@sjtu.edu.cn}{\texttt{qianc@sjtu.edu.cn}}
\end{minipage}}}

%% file: authors_pdf.tex
\author{%
\begin{minipage}[t]{\dimexpr\textwidth-2\tabcolsep\relax}
\centering\bfseries
\mbox{Lin Wu$^{1}$\hspace{8pt}Shu Yao$^{1}$\hspace{8pt}Yuheng Wang$^{1}$\hspace{8pt}Jingru Fan$^{1}$\hspace{8pt}Huatao Li$^{1}$\hspace{8pt}Yihan Li$^{1}$}\\
\mbox{Liu Yixuan$^{1}$\hspace{8pt}Cheng Liu$^{1}$\hspace{8pt}Jian Chang$^{2}$\hspace{8pt}Zeyu Zhao$^{2}$\hspace{8pt}Aohua Zhang$^{2}$\hspace{8pt}Huiying Zhao$^{2}$}\\
\mbox{Shuyuan Zhu$^{2}$\hspace{8pt}Chen Qian$^{1}$}\\[4pt]
\normalfont
\coevokeSAISymbol\hspace{5pt}$^{1}$ Shanghai Jiao Tong University\\
\coevokeUnicomSymbol\hspace{5pt}$^{2}$ China Unicom Online Information Technology CO.,Ltd\\[2pt]
{\small\ttfamily\href{mailto:linwuai2026@gmail.com}{linwuai2026@gmail.com}\quad
\href{mailto:qianc@sjtu.edu.cn}{qianc@sjtu.edu.cn}}
\end{minipage}}

%% file: preprint_title.tex
\fancypagestyle{coevokeTitle}{%
  \fancyhf{}%
  \fancyhead[L]{Preprint}%
  \fancyfoot[C]{\thepage}%
  \renewcommand{\headrulewidth}{0.4pt}%
}

%% file: sections/abstract.tex
\begin{abstract}
Training multi-agent collaboration requires trajectories of checking, feedback, and repair, yet task-solving rollouts may end in immediate approval or repeat unsuccessful attempts without useful correction. We introduce CoEvoke, a cross-domain framework that evokes these interactions during trajectory collection. Controlled perturbations create recoverable difficulties, while restricted reference-based guidance directs investigation. The resulting dataset contains 4,567 reconstructable trajectories over 4,130 software-engineering and reasoning tasks, linking public interactions to evidence, decisions, and task-native outcomes. Separate experiments over 2,287 same-task pairs show feedback increases of 47.72--100 percentage points across seven perturbation groups; guidance has smaller, model-dependent effects. With identical training tasks, decision-record counts, and updates, perturbation-enriched judge training improves ACPBench-Hard success from 35.38\% to 42.12\% (+6.73 points). SWE verifier training gains 7.60--9.40 points over interface-matched untrained references. Workflow and post-training evaluations preserve task-solving topology with unperturbed inputs and no private-reference guidance. These results support designing collection conditions to obtain corrective supervision that improves collaborative decisions with frozen solving agents.
\end{abstract}

%% file: sections/introduction.tex
\section{Introduction}
\label{sec:introduction}

Large language models (LLMs) support tool-using agents that act, observe consequences, and choose subsequent actions~\citep{liu2024agentbench,ICLR2024_8a0d3ae9}. Multi-agent systems (MAS) coordinate specialized agents through delegation, discussion, and mutual review~\citep{li2023camel,wu2024autogen,ICLR2025_66a026c0}. Success depends on exchanging evidence, evaluating intermediate work, and deciding when to revise or stop~\citep{du2023multiagentdebate,dang2025orchestration}. Learning these decisions at scale requires diverse trajectories linking actions and information exchanges to feedback, revisions, and outcomes~\citep{motwani2025malt,zeng2024agenttuning}, motivating automated collection across tasks and collaboration patterns with varying demands for checking, feedback, and repair.

However, task-solving rollouts do not guarantee trajectories for learning useful collaboration. Prior work uses agent trajectories for refinement, multi-agent training, and distillation~\citep{ICLR2025_a3cc5012,motwani2025malt,luo2026agentark,li2025agentfoundation}, but such rollouts can omit the error diagnosis, targeted feedback, and revision choices that collaborating agents need to learn. An immediately correct answer bypasses diagnosis and repair; unsuccessful teams may repeat searches and revisions without a productive direction~\citep{ICLR2026_2f5fdce2}. More rollouts therefore need not teach agents how to turn targeted feedback into useful correction.

In human collaboration, an inconsistency can evoke cross-checking, while a focused suggestion directs teammates toward useful evidence. Inspired by this pattern, we create recoverable difficulties and provide restricted guidance for investigation. These controls give agents a reason to check and a direction for repair, retaining task objectives and repair capacity.

\begin{figure}[t]
\centering
\includevectorpdf[width=\linewidth]{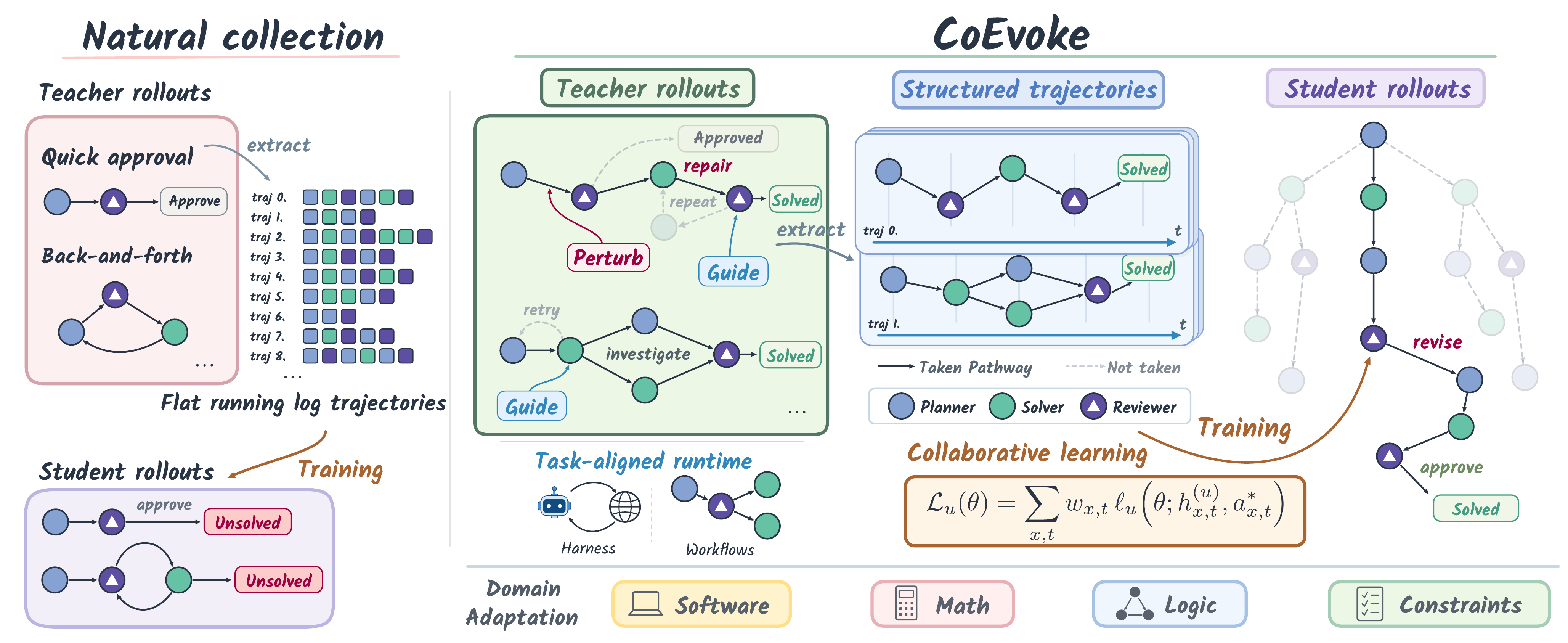}
\caption{During collection, Perturb evokes checking and repair; Guide directs investigation. Circles denote role-colored agent turns (triangle: decision); squares denote log entries. Solid/dashed paths are taken/untaken; $t$ denotes time. Stacked trajectories retain interaction links and supply public-context/target-action pairs for collaborative training (Eq.~\ref{eq:controltraining}). Student evaluation preserves workflows without perturbations or private-reference guidance.}
\label{fig:overview}
\end{figure}

We introduce \textbf{CoEvoke} (Collaboration + Evoke), an automated cross-domain framework for collecting reconstructable multi-agent trajectories. It perturbs an initial candidate, observation, or model assignment to evoke checking and repair, and provides restricted reference-based guidance to direct investigation. We call these operations \emph{Perturb-and-Guide}. The resulting interactions expose judgment, criticism, and verification for training control roles that assess intermediate work and route execution; solving agents remain frozen during this supervised training stage (Figure~\ref{fig:overview}).

We evaluate CoEvoke in four domains requiring trajectory supervision for learning collaboration: software engineering, mathematics, logical entailment, and constraint solving. Source identities, role/tool events, visibility labels, and outcome links support reconstruction and training; human-calibrated audits assess quality. Mechanism and training experiments test whether collection exposes useful correction and improves learned decisions in downstream workflows.

\noindent Our contributions are threefold:
\begin{itemize}
\item We introduce \textbf{CoEvoke}, a cross-domain framework for collecting reconstructable multi-agent trajectories through Perturb-and-Guide. It creates recoverable difficulties and directs investigation, exposing checking, feedback, and repair that task-solving rollouts may omit.
\item We construct a \textbf{dataset of 4,567 CoEvoke trajectories across 4,130 tasks}, linking public role and tool events, feedback and revisions, and evaluator outcomes. These links support inspection of the recorded public collaboration and extraction of decision records\footnote{\label{fn:record}A trajectory is one run's retained public event sequence, with messages, tool observations, and outcome links. It can yield several decision records, each pairing the public context available at one decision with its target action; later repairs and evaluator outcomes are excluded from that input. Trajectory and record counts therefore need not match.} for training control roles to decide between candidate acceptance and further repair.
\item We demonstrate the \textbf{training utility of CoEvoke trajectories with frozen solvers}. Matched-data judge training improves ACPBench-Hard success by 6.73 percentage points; interface-matched verifier training gains 7.60--9.40 points on SWE-bench Verified. Evaluation uses unperturbed inputs without private-reference guidance.
\end{itemize}

%% file: sections/related_work.tex
\section{Related Work}
\label{sec:related}

\subsection{From Reasoning to Coordinated Agents}
Reasoning extends intermediate-step generation~\citep{wei2022chainofthought} through candidate aggregation~\citep{wang2023selfconsistency} and tool interaction~\citep{yao2023react,schick2023toolformer}, whose evidence interfaces affect execution~\citep{yang2024sweagent}. CAMEL studies role-playing cooperation, while AutoGen, MetaGPT, and ChatDev organize communication and task decomposition~\citep{li2023camel,wu2024autogen,hong2024metagpt,qian2024chatdev}. Debate, scalable collaboration graphs, and learned orchestration address candidate comparison and agent sequencing~\citep{du2023multiagentdebate,ICLR2025_66a026c0,dang2025orchestration}. CoEvoke studies how collection conditions in these workflows evoke checking and repair.

\subsection{Synthetic Data and Agent Trajectory Generation}
As synthetic supervision moves from instructions and rationales to executable interactions~\citep{wang2023selfinstruct,zelikman2022star,zeng2024agenttuning}, collection becomes a design problem. AgentTrek uses tutorials to guide execution, APIGen-MT builds verifiable task blueprints, and SWE-smith generates repository tasks through test-breaking changes~\citep{xu2025agenttrek,prabhakar2025apigenmt,yang2025swesmith}. Agent Data Protocol complements generation by unifying heterogeneous trajectories for fine-tuning~\citep{ICLR2026_e79c09f3}.

AgentArk and Agent Foundation Models distill team behavior into individual models~\citep{luo2026agentark,li2025agentfoundation}. AgentRefine synthesizes environments for corrective supervision~\citep{ICLR2025_a3cc5012}, while MALT trains role-specific models through generator--verifier--refiner trees and propagated rewards~\citep{motwani2025malt}; CoEvoke instead intervenes on existing tasks to collect trajectories for training control roles with frozen solvers.

\subsection{Feedback, Verification, and Process Evaluation}
Self-refinement, remembered reflections, and tool-assisted critique supply feedback~\citep{madaan2023selfrefine,shinn2023reflexion,gou2024critic}, but unaided self-correction can fail~\citep{huang2024selfcorrect,ICLR2025_f3c5e562}; SCoRe learns correction through on-policy interaction~\citep{kumar2025score}. Learned verifiers and process supervision distinguish answer quality from reasoning quality~\citep{cobbe2021verifiers,uesato2022process,lightman2023verifystep}, motivating separate measures of multi-turn reliability and coordination~\citep{laban2026lost,zhu2025multiagentbench}.

AgenTracer injects faults for attribution training, whereas DoVer tests debugging hypotheses by re-execution~\citep{ICLR2026_134ed7a4,ICLR2026_2f5fdce2}. CoEvoke uses interventions to collect feedback-and-repair interactions, assessing outcomes with task-native evaluators~\citep{jimenez2024swebench,ICLR2024_8a0d3ae9,ICLR2025_1b126cc3} and process quality with human-calibrated audits~\citep{zheng2023llmjudge,liu2023geval,kim2024prometheus}. Agent approval and task success remain distinct evidence.

%% file: sections/method.tex
\section{CoEvoke: Evoking Collaboration during Collection}
\label{sec:coevoke}

\begin{figure}[t]
\centering
\includevectorpdf[width=\linewidth]{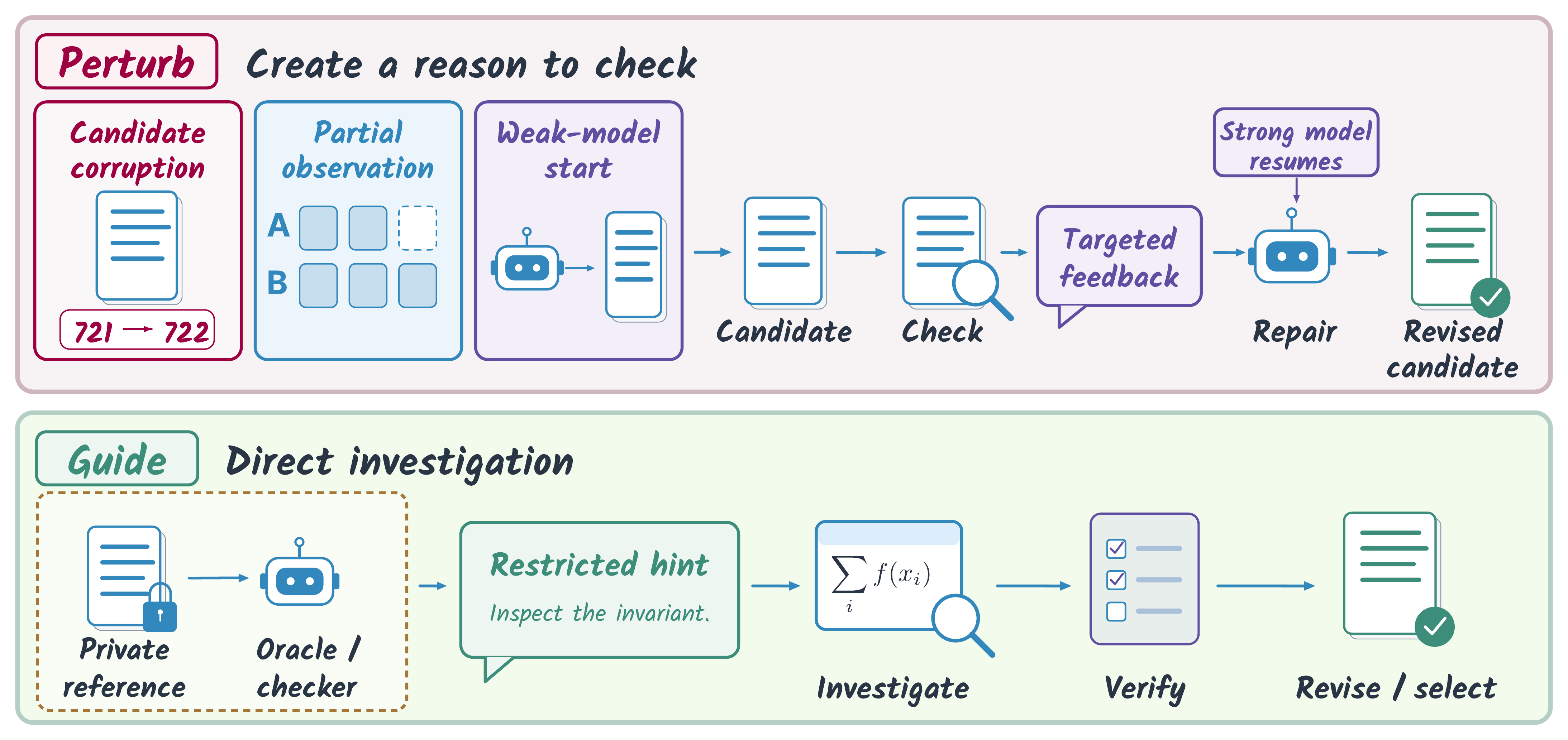}
\caption{Complementary collection controls. Top: candidate corruption, partial observation, or weak initialization evokes checking and repair; the strong model performs requested repairs. Bottom: an oracle or checker converts private references into restricted hints for public investigation and verification. Task objectives and topology are preserved. Icons and values are illustrative; outcomes use task-native evaluation.}
\label{fig:method}
\end{figure}

To collect trajectories that expose decisions absent from natural execution, CoEvoke creates recoverable difficulties and directs investigation. Figure~\ref{fig:method} summarizes the design: retain the task-solving workflow, adjust its collection conditions, and preserve evidence-to-decision links for training.

\subsection{Tasks and Workflows}
\label{sec:workflows}
Useful correction must remain anchored to an independently checkable task. We therefore represent a task as $x=(c_x,z_x,E_x)$, with public context $c_x$, optional private references $z_x$, and a task-native evaluator $E_x$. Its workflow $G_x=(V_x,A_x)$ specifies roles and permitted communication routes. Only a designated collection node may read $z_x$; an existing critic or judge can serve this role without adding another agent.

Different uncertainties call for different coordination patterns~\citep{chen2023agentverse,motwani2025malt,du2023multiagentdebate}. Software repair uses planning, coding, and review on SWE-smith~\citep{yang2025swesmith}; mathematical reasoning returns DeepMath critiques to a solver~\citep{he2025deepmath}; candidate comparison resolves ProofWriter and Zebra disagreements through independent debaters, a judge, and targeted revision~\citep{tafjord2021proofwriter,lin2025zebralogic}. In every workflow, task-native evaluation independently scores the final solution.

To prevent task overlap, all variants share a CoEvoke split, and external benchmarks are excluded from construction and training. Downstream evaluation retains task-solving roles and routing, with unperturbed inputs and no private-reference guidance (Section~\ref{sec:experiments}).

\subsection{Perturb-and-Guide}
\label{sec:intervention_learning}
Corrective collaboration needs both a reason to question an intermediate result and a feasible route to repair it. Perturb creates recoverable discrepancies, while Guide provides restricted direction for investigation. For role $v$ with public input $h_t^v$ and assigned model $m_t^v$ at step $t$, a single collection operation is
\begin{equation}
\mathcal C_\omega(h_t^v,m_t^v,z_x)=
\begin{cases}
(h_t^v,m_t^v), & \omega=\mathrm{natural},\\
I_\kappa(h_t^v,m_t^v), & \omega=\mathrm{perturb},\\
(h_t^v\oplus\Pi_\rho(h_t^v,z_x),m_t^v), & \omega=\mathrm{guide}.
\end{cases}
\label{eq:intervention}
\end{equation}
Here $I_\kappa$ perturbs an input or initial model assignment, $\Pi_\rho$ restricts reference-informed assistance, and $\oplus$ delivers it to the permitted role. Operations can be combined at designated roles; controlled experiments isolate their effects.

\paragraph{Perturb: create a reason to check.}
The runtime perturbs only a designated initial interaction, preserving evidence and repair capacity:
\begin{equation}
\begin{aligned}
\text{Candidate corruption:}\quad & \tilde y_0=\kappa_x(y_0),\quad q_x(y_0)=1,\quad q_x(\tilde y_0)=0;\\
\text{Partial observation:}\quad & \tilde F_x^{\,v}=F_x\setminus\{f\};\\
\text{Weak initialization:}\quad & m_{\mathrm{init}}^v=m_{\mathrm{weak}},\quad m_{\mathrm{repair}}^v=m_{\mathrm{strong}}.
\end{aligned}
\label{eq:perturb_operations}
\end{equation}
Here $y_0$ is the first candidate and $q_x$ a deterministic task-specific correctness check; $\kappa_x$ changes an integer, entailment label, or table value. Both checks must hold; otherwise $y_0$ is retained. For partial observation, $F_x$ is the public fact set and $f$ a uniquely occurring fact on an unambiguous ProofWriter proof path. Other debaters and the judge retain all facts to diagnose the missing premise. Weak initialization restores the original stronger model only when feedback requests repair. No operation is reapplied during repair; approval ends checking, and revisions obey shared budgets.

\paragraph{Guide: direct investigation.}
A recoverable difficulty still needs an actionable direction. Guide maps the current public context and private reference to restricted assistance for a permitted recipient:
\begin{equation}
g_t^v=\Pi_\rho(h_t^v,z_x),\qquad h_t^{v,+}=h_t^v\oplus g_t^v.
\label{eq:guide_delivery}
\end{equation}
Here $\rho$ specifies the recipient, structure, budget, and content restrictions; $g_t^v$ is the guidance and $h_t^{v,+}$ the augmented public input. In software engineering, an oracle sends hints to the planner; in reasoning, the existing critic or judge returns feedback. Public solvers construct and verify solutions. Runtime controls isolate private inputs, enforce recipients and structure, and reject known direct-answer and private-source patterns in reasoning feedback. Prompts prohibit target patches and complete solutions. These constraints do not guarantee semantic nondisclosure, so guidance is audited separately (Section~\ref{sec:dataset}).

\subsection{From Reconstructable Trajectories to Decision Supervision}
\label{sec:eventgraph}
To reuse corrective interactions, we preserve retained public evidence and outcomes in a temporal event graph $\mathcal T_x=(\mathcal N_x,\mathcal R_x,\prec,\nu,\phi)$. Nodes $\mathcal N_x$ are role/tool events; relations $\mathcal R_x$ link messages, feedback, revisions, selections, and tool responses; $\prec$ records order, $\nu$ visibility, and $\phi$ the final source. Reconstruction covers this exported interaction, not every teacher-visible input: immediate SWE test-pass summaries may influence collection-time review but enter neither public trajectories nor training contexts.

The full recorded trajectory supplies decision-specific supervision; it need not be supplied in full to each controller. For control role $u$, $h_{x,t}^{(u)}=P_u(\mathcal T_{x,\leq t})$ extracts the retained public context before its target action. Targets specify selection or repair for the judge, feedback for the critic, and acceptance or rejection for the verifier. Inputs exclude private references, subsequent revisions, and evaluator outcomes. The training objective is
\begin{equation}
\mathcal L_u(\theta)=\sum_{x,t}w_{x,t}\,
\ell_u\!\left(\theta;h_{x,t}^{(u)},a_{x,t}^{*}\right),
\label{eq:controltraining}
\end{equation}
where $a_{x,t}^{*}$ is the target decision, $\sum_t w_{x,t}=1$ normalizes each task's contribution, and $\ell_u$ is weighted token-level negative log-likelihood. Task normalization prevents long trajectories from dominating. Frozen actors isolate learning of specific decisions (Section~\ref{sec:learning_results}).

%% file: sections/dataset.tex
\section{Data Generated by CoEvoke}
\label{sec:dataset}

\subsection{Scale and Composition}
Repository repair requires long-horizon tool use and executable verification~\citep{jimenez2024swebench,yang2024sweagent}; reasoning exposes derivation errors, disagreement, and constraint violations. Across these settings, the dataset contains 4,567 reconstructable trajectories over 4,130 tasks (Table~\ref{tab:scale}). Inclusion requires complete public role outputs and interaction links, irrespective of task success. Source tasks, role events, revisions, and evaluated artifacts remain linked.

\begin{table}[t]
\setlength{\belowcaptionskip}{7pt}
\caption{\textbf{The dataset includes only reconstructable trajectories.} Experimental runs are excluded even where source tasks overlap; a task can contribute multiple collection trajectories. Outcomes come from task-native evaluators.}
\label{tab:scale}
\centering
\small
\setlength{\tabcolsep}{5pt}
\renewcommand{\arraystretch}{1.10}
\begin{tabularx}{\linewidth}{YCCCC}
\toprule
\rowcolor{tblHead}
\tblhead{Source} & \tblhead{Tasks} & \tblhead{Trajectories} & \tblhead{\pass} & \tblhead{\fail} \\
\midrule
SWE-smith & 1,763 & 1,763 & 1,561 & 202 \\
\rowcolor{tblStripe}
DeepMath & 831 & 923 & 917 & 6 \\
ProofWriter & 1,096 & 1,392 & 1,392 & 0 \\
\rowcolor{tblStripe}
Zebra & 440 & 489 & 489 & 0 \\
\midrule
\rowcolor{tblSummary}
\textbf{Total} & \textbf{4,130} & \textbf{4,567} & \textbf{4,359} & \textbf{208} \\
\bottomrule
\end{tabularx}
\end{table}

Validation checks identifiers, chronology, call--observation pairs, feedback/revision references, final sources, and evaluator links. Public exports omit credentials, internal paths, private references, hidden reasoning, and immediate SWE evaluation counts. These counts may affect collection-time review but enter no training context; reconstruction covers retained public events. Distinct task, trajectory, and communication identities support pairing each public prefix with its target action for training.

\subsection{Human and Model Audits}
Because task success alone does not establish collaboration quality, we assess seven dimensions of contribution, evidence use, coherence, and efficiency. Grades A--D summarize process quality alongside task-native outcomes; a separate scale distinguishes no observed disclosure (L0), references to private authority (L1), protected-content disclosure (L2), and insufficient context (LX). Five experts contribute 240 ratings over 160 trajectories to calibrate model assessment and evidence-based adjudication~\citep{zheng2023llmjudge,liu2023geval,kim2024prometheus}.

\begin{table}[t]
\setlength{\belowcaptionskip}{5pt}
\caption{\textbf{Human--LLM agreement and trajectory quality.} Agreement pools five experts; quality and disclosure proportions use only the stratified probability sample.}
\label{tab:audit}
\centering
\small
\setlength{\tabcolsep}{3.5pt}
\renewcommand{\arraystretch}{1.08}
\begin{tabularx}{\linewidth}{>{\hsize=1.2\hsize}Y>{\hsize=1.4\hsize}C>{\hsize=.65\hsize}C>{\hsize=.75\hsize}C}
\toprule
\rowcolor{tblSection}
\multicolumn{4}{l}{\tblhead{\textcolor{tblInk}{\textbf{A. Human--LLM agreement}}\quad\textit{240 ratings; 160 trajectories}}} \\
\rowcolor{tblHead}
\tblhead{Dimension} & \tblhead{Exact agreement $\uparrow$} & \tblhead{$\kappa_w$ / $\kappa$} & \tblhead{Score MAE $\downarrow$} \\
\midrule
Collaboration grade & \tblkey{201/240 (83.8\%)} & 0.779 & 1.10 \\
\rowcolor{tblStripe}
Disclosure label & \tblkey{218/240 (90.8\%)} & 0.688 & -- \\
\end{tabularx}
\par\nointerlineskip
\begin{tabularx}{\linewidth}{L{.21\linewidth}CCCCCCCC}
\midrule
\rowcolor{tblSection}
\multicolumn{9}{l}{\tblhead{\textcolor{tblInk}{\textbf{B. Probability-sample audit}}\quad\textit{Design-weighted proportions (\%)}}} \\
\iflatexml
% HTML row spans start in the first row; the PDF retains its color-safe layout.
\rowcolor{tblHead}
\multirow{2}{*}{\tblhead{Dataset}} & \multirow{2}{*}{\tblhead{$n$}} & \multicolumn{4}{c}{\tblhead{Collaboration grade}} & \multicolumn{3}{c}{\tblhead{Disclosure label}} \\
\rowcolor{tblHead}
 & & \tblhead{A} & \tblhead{B} & \tblhead{C} & \tblhead{D} & \tblhead{L0} & \tblhead{L1} & \tblhead{L2} \\
\else
\rowcolor{tblHead}
 &  & \multicolumn{4}{c}{\tblhead{Collaboration grade}} & \multicolumn{3}{c}{\tblhead{Disclosure label}} \\

\rowcolor{tblHead}
\multirow{-2}{*}{\tblhead{Dataset}} & \multirow{-2}{*}{\tblhead{$n$}} & \tblhead{A} & \tblhead{B} & \tblhead{C} & \tblhead{D} & \tblhead{L0} & \tblhead{L1} & \tblhead{L2} \\
\fi
\midrule
SWE Oracle & 205 & 85.1 & 0.0 & 14.9 & 0.0 & 96.1 & 3.9 & 0.0 \\
\rowcolor{tblStripe}
DeepMath & 97 & 97.9 & 1.0 & 1.0 & 0.0 & 97.9 & 2.1 & 0.0 \\
ProofWriter & 146 & 98.6 & 1.4 & 0.0 & 0.0 & 100.0 & 0.0 & 0.0 \\
\rowcolor{tblStripe}
Zebra & 52 & 100.0 & 0.0 & 0.0 & 0.0 & 100.0 & 0.0 & 0.0 \\
\midrule
\rowcolor{tblSummary}
\textbf{Overall} & \textbf{500} & \textbf{93.1} & \textbf{0.6} & \textbf{6.3} & \textbf{0.0} & \textbf{98.0} & \textbf{2.0} & \textbf{0.0} \\
\bottomrule
\end{tabularx}
\par\vspace{5pt}
\begin{minipage}{\linewidth}\footnotesize
Grades use linearly weighted $\kappa_w$ and a 0--14 score; disclosure uses unweighted $\kappa$. In B, $n$ counts trajectories, not weighted observations. L1 denotes private-authority disclosure and L2 protected-content disclosure; no LX labels were observed.
\end{minipage}
\end{table}

Table~\ref{tab:audit} separates human agreement from quality estimation: pooled agreement includes probability-sampled and targeted cases, while quality estimates use the weighted 500-trajectory probability sample. Grade A accounts for 93.1\%; 2.0\% is labeled L1, and no L2 disclosure is detected. Agreement supports label reliability, while the grade distribution indicates generally high-quality recorded interaction. Coupled with linked public role/tool evidence and reconstruction and visibility checks, these results support an information-rich corpus ready for decision-specific training. These weighted estimates describe the fixed probability-audit design, not a census of the corpus; zero observed L2 cases does not guarantee zero population risk.

%% file: sections/experiments.tex
\section{Experiments}
\label{sec:experiments}

\subsection{Evaluation Design}
Only collection-mechanism experiments activate Perturb or Guide. Workflow and post-training evaluations retain task-solving topology, ordinary feedback, and repair, using unperturbed inputs and public task/execution evidence without private-reference guidance. Hosted teachers test collection; Qwen3.5-9B, Gemma-4-12B-IT, and Ministral-3-8B-Instruct test downstream use with fixed solving weights, decoding, and output protocols, with explicit thinking disabled.

\paragraph{Collection controls.}
The perturbation experiment contains 2,287 same-task pairs: 1,197 compare original and perturbed inputs, and 1,090 compare strong-first and weak-to-strong generation. Each pair fixes the checker (DeepSeek-V4-Pro, medium thinking), reference access, and evaluator; weak initialization uses Qwen2.5-7B-Instruct. Guidance compares reference-based with public-only assistance: the same 80 SWE-smith tasks across five models give 400 runs per condition; reasoning uses 240 assigned tasks per model. These experimental trajectories are excluded from the dataset.

\paragraph{Downstream evaluation.}
Workflow tests use \emph{Single-MA}, one agent with the multi-agent workflow's aggregate model-call, tool-use, and token allowances. Primary evaluations cover all 500 SWE-bench Verified and 1,040 ACPBench-Hard tasks and a fixed 144-task MathArena collection~\citep{jimenez2024swebench,balunovic2025matharena,kokel2026acpbenchhard}; internal 100-task samples supplement them. Training updates only the judge, critic, or verifier. Natural-SFT and Perturb-SFT match training task IDs, decision records, and updates; each verifier is compared with an untrained model receiving the same evidence format.

\paragraph{Measures.}
Task-native success uses prespecified denominators. Process measures distinguish a feedback message, an executed revision, an evaluator-approved revision, and its selection as the final source. A selected correct revision additionally requires an answer change. We report percentage-point differences and paired intervals or tests where supported; the fixed-sample SWE guidance comparison is descriptive. Reasoning intervals use task-paired bootstrap resampling over evaluable pairs. Full-benchmark success retains all assigned tasks in its denominator, including unavailable outcomes. SWE net provider tokens sum reported per-call usage after excluding explicit failed outputs and schema/API retries; ordinary investigation, test failures, feedback, revision, and guidance calls remain included.

%% file: sections/results.tex
\label{sec:results}

\subsection{Do Perturbations Evoke Corrective Interaction?}
\label{sec:intervention_results}
\begin{figure}[t]
\centering
\includegraphics[width=\linewidth]{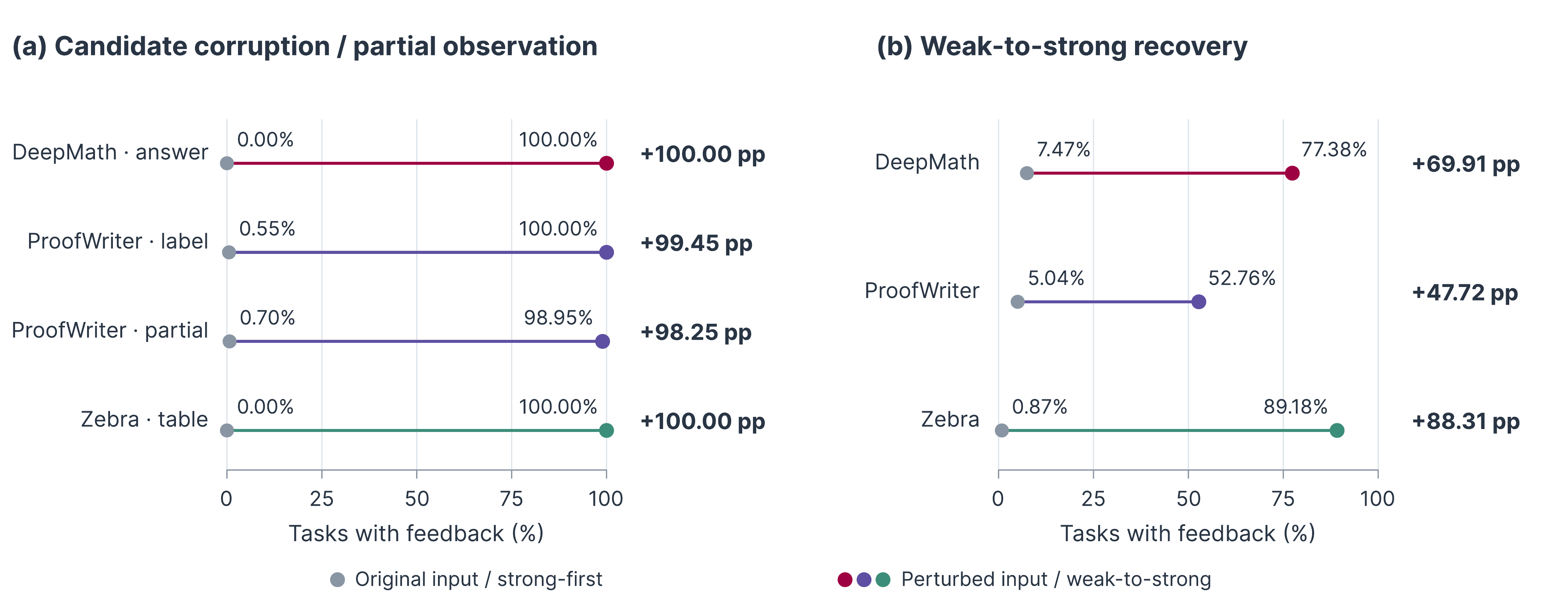}
\caption{Perturb increases feedback in (a) 1,197 candidate-corruption/partial-observation pairs and (b) 1,090 strong-first/weak-to-strong pairs. Markers connect control (gray) and perturbed rates; colors identify datasets and labels give percentage-point gains. All pairs are evaluable. We use task-paired bootstrap resampling for 95\% confidence intervals.}
\label{fig:intervention}
\end{figure}

For 1,197 input-perturbation pairs, feedback rises from 0.33\% to 99.75\%, a 99.42-point increase (95\% CI 98.91--99.83; Figure~\ref{fig:intervention}). The runtime assigns a weaker model only to initial generation; across 1,090 weak-to-strong pairs, feedback rises from 5.14\% to 70.46\%, or 65.32 points (95\% CI 62.57--68.07).

Both conditions solve all 1,197 input-perturbation tasks, but 732 perturbed runs (61.15\%) yield a \emph{selected correct revision}.\footnote{\label{fn:revision}A run is counted once when its final source is a revision, its normalized answer differs from that role's first candidate delivered for checking in the same run, and the final task outcome is \pass{}. For candidate corruption, the comparison starts from the perturbed answer.} This is not an overall repair-success rate: the other 465 runs (38.85\%) also pass, without satisfying that joint criterion. All 344 DeepMath runs select a revised final answer; multi-candidate workflows may generate a correct revision yet select another valid candidate. Perturb thus supplies corrective supervision even when final accuracy cannot improve.

Under weak initialization, final success remains unchanged on ProofWriter and Zebra and moves from 87.78\% to 86.65\% on DeepMath, with a paired interval spanning zero. The additional interaction costs 1.87 model calls per task, versus 2.31 for input perturbations, supplying corrective supervision without guaranteeing more solved tasks.

\subsection{Does Restricted Guidance Improve Task Solving?}
\label{sec:information_results}
The public-only control O1 matches reference-based O2's planner-facing schema, guidance role, and output budget. Across the same 80 SWE-smith tasks and five models, O2 raises success from 348/400 (87.00\%) to 361/400 (90.25\%), a 3.25-point gain (Table~\ref{tab:oracle}). Four models improve and one ties; all five use fewer net provider tokens, with an aggregate reduction of 8.55\%.

\begin{table}[t]
\setlength{\belowcaptionskip}{5pt}
\caption{\textbf{Model-dependent effects of reference access.} (A) The same 80 SWE tasks $\times$ five models give 400 runs per arm. Net provider-token change is $100(T_{\mathrm{O2}}/T_{\mathrm{O1}}-1)\%$ (lower is better); the total uses summed tokens. (B) Reasoning rates share $n$ paired tasks. O1 is public-only and O2 reference-based guidance. Colors highlight gains and token reductions; preferred directions appear in headers.}
\label{tab:oracle}
\centering
\footnotesize
\renewcommand{\arraystretch}{1.04}
\setlength{\tabcolsep}{3.5pt}
\begin{tabularx}{\linewidth}{Yccccc}
\toprule
\rowcolor{tblSection}
\multicolumn{6}{l}{\tblhead{\textcolor{tblInk}{\textbf{A. SWE: Oracle guidance to the Planner}}}} \\
\rowcolor{tblHead}
\tblhead{Model (thinking)} & \tblhead{$n$} & \tblhead{\tblstack{O1\\(\%)}} & \tblhead{\tblstack{O2\\(\%)}} & \tblhead{\tblstack{Gain\\(pp) $\uparrow$}} & \tblhead{\tblstack{Net provider\\token change (\%) $\downarrow$}} \\
\midrule
GPT-5.4 (medium) & 80 & 77.50 & 83.75 & \textcolor{tblRise}{+6.25} & \textcolor{tblTokenBlue}{-9.97} \\
\rowcolor{tblStripe}
Claude Sonnet 5 (adaptive; medium) & 80 & 88.75 & 88.75 & \same{+0.00} & \textcolor{tblTokenBlue}{-17.95} \\
DeepSeek-V4-Pro (enabled; medium) & 80 & 88.75 & 90.00 & \textcolor{tblRise}{+1.25} & \textcolor{tblTokenBlue}{-9.54} \\
\rowcolor{tblStripe}
GLM-5.2 (enabled; max) & 80 & 92.50 & 96.25 & \textcolor{tblRise}{+3.75} & \textcolor{tblTokenBlue}{-4.19} \\
Kimi K2.6 (enabled; provider default) & 80 & 87.50 & 92.50 & \textcolor{tblRise}{+5.00} & \textcolor{tblTokenBlue}{-3.91} \\
\midrule
\rowcolor{tblSummary}
\textbf{Total} & 400 & 87.00 & \textbf{90.25} & \textcolor{tblRise}{+3.25} & \textcolor{tblTokenBlue}{-8.55} \\
\midrule
\rowcolor{tblSection}
\multicolumn{6}{l}{\tblhead{\textcolor{tblInk}{\textbf{B. Reasoning: reference-informed Critic/Judge}}}} \\
\rowcolor{tblHead}
\tblhead{Model (thinking)} & \tblhead{$n$} & \tblhead{Public} & \tblhead{Reference} & \tblhead{\tblstack{$\Delta$ success\\(pp) $\uparrow$}} & \tblhead{\tblstack{Paired 95\% CI\\(pp)}} \\
\midrule
GPT-5.4 (medium) & 240 & 81.25 & 85.00 & \textcolor{tblRise}{+3.75} & [1.67, 6.25] \\
\rowcolor{tblStripe}
Claude Sonnet 5 (adaptive; medium) & 236 & 78.81 & 81.78 & \textcolor{tblRise}{+2.97} & [0.85, 5.08] \\
DeepSeek-V4-Pro (enabled; medium) & 236 & 92.37 & 94.92 & \textcolor{tblRise}{+2.54} & [0.42, 5.08] \\
\rowcolor{tblStripe}
GLM-5.2 (enabled; max) & 237 & 85.65 & 87.34 & \textcolor{tblRise}{+1.69} & [0.42, 3.80] \\
Kimi K2.6 (enabled; provider default) & 234 & 92.74 & 93.59 & \textcolor{tblRise}{+0.85} & [0.00, 2.14] \\
\bottomrule
\end{tabularx}
\end{table}

Reference-informed reasoning gains range from 0.85 to 3.75 points across all five models, with limited headroom where public checking already succeeds. SWE token reductions range from 3.91\% to 17.95\%; Claude preserves success while reducing usage. These descriptive provider-token totals do not represent a common tokenizer or monetary cost.

\subsection{Are the Workflows Effective without Perturb-and-Guide?}
\label{sec:system_results}
\begin{figure}[t]
\centering
\includegraphics[width=\linewidth]{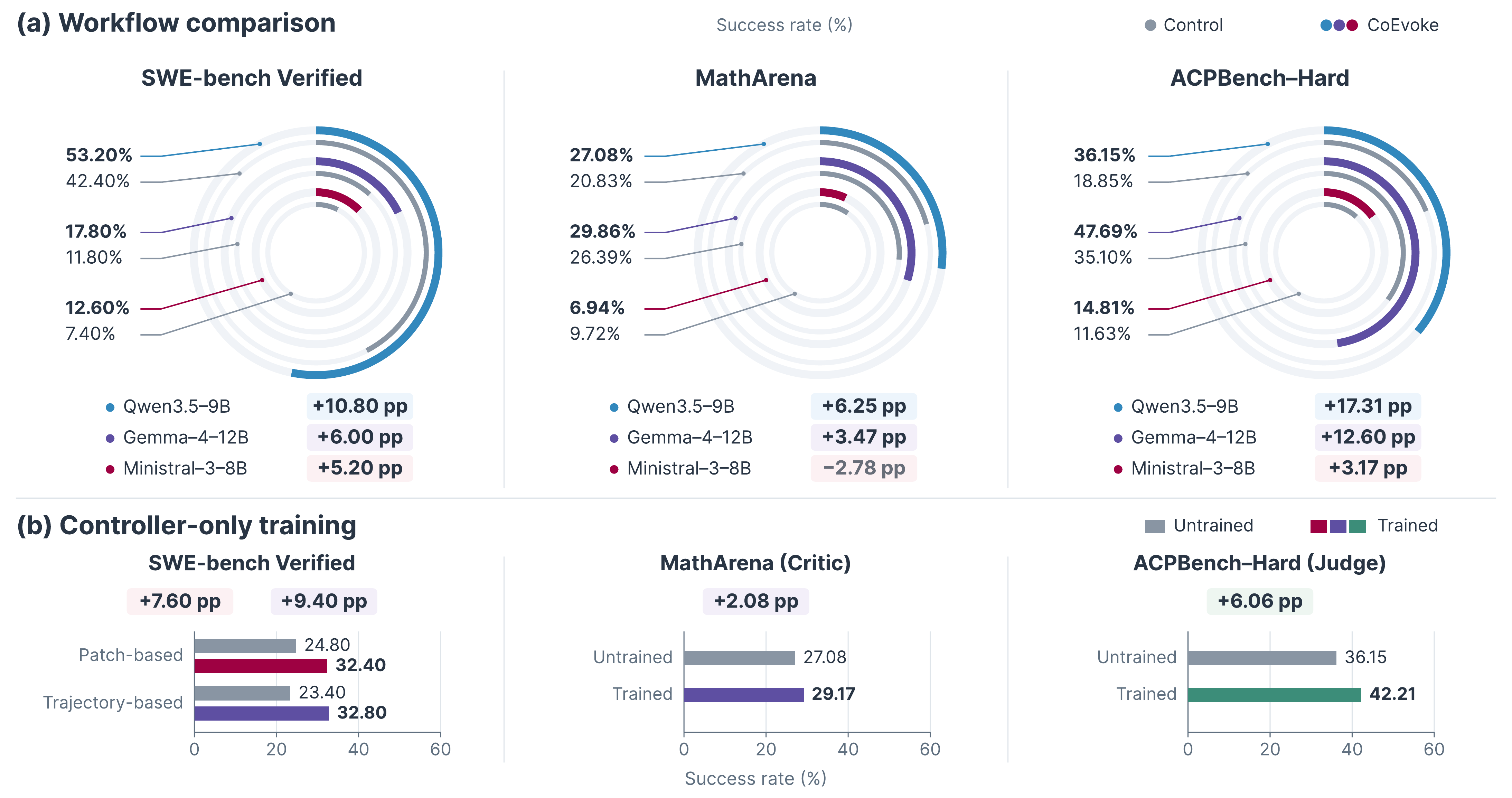}
\caption{Workflow and training comparisons. (a) Shared-budget control (gray) versus CoEvoke; circles represent 100\%, with Qwen, Gemma, and Ministral outside inward. (b) Control-role training with frozen Qwen solvers: the separate full-corpus judge study and interface-matched verifiers. In SWE, the Reviewer in (a) reads files and runs public tests; both Verifiers in (b) have no tools. Denominators: 500 SWE-bench Verified, 144 MathArena, 1,040 ACPBench-Hard. Evaluation uses unperturbed inputs, no private-reference guidance, and no explicit thinking.}
\label{fig:system}
\label{fig:learning}
\end{figure}

On unperturbed inputs without private-reference guidance, CoEvoke exceeds budget-matched single-agent controls in eight of nine external model--benchmark combinations (Figure~\ref{fig:system}a). All three models gain on SWE-bench Verified and ACPBench-Hard. On MathArena, Qwen and Gemma gain 6.25 and 3.47 points, while Ministral loses 2.78 points.

Qwen's internal controls solve 87\% of DeepMath and 84\% of ProofWriter; CoEvoke loses one point on each, whereas Zebra rises from 52\% to 60\%. Benefits vary by task and model, supporting workflow effectiveness, not topology optimality.

\subsection{Does Training on CoEvoke Trajectories Improve Task Solving?}
\label{sec:learning_results}
With 884 shared tasks, 891 decision records, and 379 updates, Perturb-SFT solves 438/1,040 ACPBench-Hard tasks (42.12\%) versus Natural-SFT's 368/1,040 (35.38\%), a 6.73-point gain. Repair-routing targets grow from 7 (0.79\%) to 440 (49.38\%): the fixed budget shifts from almost exclusively approval targets to substantial repair-routing supervision, realizing the intended collection effect. Natural-SFT uses independently collected clean trajectories; both sources use reference-informed checking during collection. This comparison uses Qwen3.5-9B at the final step-379 checkpoint and one training seed; matching records and updates does not imply equal token exposure.

On SWE-bench Verified, patch-based verifier training improves success from 24.80\% to 32.40\%; trajectory-based training improves it from 23.40\% to 32.80\%. Both evidence interfaces support improved closed-loop task success after verifier training, with solving actors and the evaluation protocol held fixed. These results establish learnability within each interface, not an advantage of longer inputs or a replacement for the tool-using reviewer (53.20\%).\footnote{\label{fn:interface}The patch interface supplies the issue and patch; the trajectory interface adds selected prior public execution. Neither verifier has tools. Their accept/reject decisions affect subsequent execution, so candidates and evidence can differ across runs: fixed interfaces mean consistent evidence-selection rules, not identical dynamic inputs or classification of a fixed final-patch set.}

%% file: sections/discussion.tex
\section{Discussion and Limitations}
\label{sec:discussion}
A matched 180-task $2\times2$ reasoning experiment supports joint corrective supervision and token savings without establishing a superadditive interaction; it covers candidate corruption with one checker. 

We validate CoEvoke on small multi-agent teams; its effectiveness in much larger systems with more complex collaboration remains untested. Training is restricted to SFT to isolate the dataset's contribution, since subsequent reinforcement learning (RL) could introduce gains from additional interaction and rewards. Whether post-SFT RL can consolidate and generalize the learned behavior, and whether the dataset improves broader agentic capabilities beyond the evaluated control decisions, remain open questions.

\section{Conclusion}
\label{sec:conclusion}
CoEvoke makes corrective collaboration an objective of trajectory collection. Recoverable perturbations and restricted guidance produce reconstructable trajectories that preserve public evidence and outcome links for decision-record extraction. Full trajectories support decision-specific supervision without requiring every controller to receive the entire interaction. Gains with frozen solvers on unperturbed inputs without private-reference guidance support collection design as a means to improve learned checking, revision, and acceptance.
\par\label{sec:main_end}

%% file: sections/statements.tex
\section*{Reproducibility Statement}
The collection controls and public-record visibility boundaries are described in Section~\ref{sec:coevoke}, dataset construction and auditing in Section~\ref{sec:dataset}, and experimental comparisons in Section~\ref{sec:experiments}. Decision supervision pairs each retained public context with its target action, excluding private references, subsequent revisions, and evaluator outcomes from the input. Workflow and post-training evaluations retain task-solving topology with unperturbed inputs and no private-reference guidance. Upstream task text, code, and patches retain their respective licenses.

\paragraph{AI assistance.}
Generative AI assisted manuscript writing, translation, and polishing; its use in trajectory generation and auditing is described in Sections~\ref{sec:coevoke} and~\ref{sec:dataset}.